\documentclass[tinyml]{acmart}

\AtBeginDocument{%
  \providecommand\BibTeX{{%
    \normalfont B\kern-0.5em{\scshape i\kern-0.25em b}\kern-0.8em\TeX}}}

\usepackage{xcolor}
\newcommand{\mv}[1]{{\color{green}#1}}

\setcopyright{rightsretained}
\copyrightyear{2022}
\acmYear{2022}

\usepackage{subfig}
\begin{document}

\title[Delta Keyword Transformer]{Delta Keyword Transformer: Bringing Transformers to the Edge through Dynamically Pruned Multi-Head Self-Attention}

\author{Zuzana Jel\v cicov\'a}
\affiliation{%
  \institution{Technical University of Denmark; and Demant A/S}
  \city{Kongens Lyngby; and Smørum}
  \country{Denmark}
  \postcode{2800}}
\email{zuje@dtu.dk}

\author{Marian Verhelst}
\affiliation{%
  \institution{MICAS, KU Leuven}
  \city{Leuven}
  \country{Belgium}
  \postcode{3000}}
\email{marian.verhelst@kuleuven.be}


\renewcommand{\shortauthors}{Jel\v cicov\'a and Verhelst}

\begin{abstract}
Multi-head self-attention forms the core of Transformer networks. However, their quadratically growing complexity with respect to the input sequence length impedes their deployment on resource-constrained edge devices. We address this challenge by proposing a dynamic pruning method, which exploits the temporal stability of data across tokens to reduce inference cost. The threshold-based method only retains significant differences between the subsequent tokens, effectively reducing the number of multiply-accumulates, as well as the internal tensor data sizes. The approach is evaluated on the Google Speech Commands Dataset for keyword spotting, and the performance is compared against the baseline Keyword Transformer. Our experiments show that we can reduce $\sim80\%$ of operations while maintaining the original $98.4\%$ accuracy. Moreover, a reduction of $\sim87-94\%$ operations can be achieved when only degrading the accuracy by 1-4\%, speeding up the multi-head self-attention inference by a factor of $\sim7.5-16$. 
\end{abstract}

\keywords{Transformers, delta computations, pruning, compression, keyword spotting, edge devices}

\maketitle

\section{Introduction}
The Transformer architecture \cite{attentionisallyouneed} is an emerging type of neural networks that has already proven to be successful in many different areas such as natural language processing \cite{bert, roberta, fewshotNLP, radford2019language}, computer vision \cite{imageWorth16x16words, imageDistillation, vitImageNET, videoTransformer}, and speech recognition \cite{conformerSpeech, transducerSpeech, multichannel, teraSpeech}. Its success lies in the multi-head self-attention (MHSA), which is a collection of attention mechanisms executed in parallel. Although Transformers achieve state-of-the-art results, deployment to resource-constrained devices is challenging due to their large size and computational complexity that grows quadratically with respect to the sequence length. Hence, self-attention, despite being extremely efficient and powerful, can easily become a bottleneck in these models. A widely used compression technique to reduce the size and computations of DNNs is pruning, that has been extensively researched throughout the years \cite{songhan2015, songhan2016, lotteryticket, structuredPruning}. An increasing number of works focusing on MHSA pruning recently emerge. These mainly aim for reducing the number of attention heads in each Transformer layer \cite{michel, voita, mccarley}, and token pruning \cite{kim2021learned, powerbert, lat, spatten}. Eliminating attention heads completely to speed up the processing might significantly impact accuracy. Therefore, token (a vector in the sequence) pruning represents a more suitable approach, where attention heads are preserved and only unnecessary tokens within the individual heads are removed.
However, most of the methods above i) require demanding training procedures that hinder utilizing a single method across various models and applications without unnecessary overhead, and ii) focus on coarse-grained pruning. \\ 
In this work, we further push pruning to finer granularity, where individual features within tokens are discarded at runtime using a threshold in the MHSA pipeline. The reduction is based on the comparison of similarities between corresponding features of subsequent tokens, where only the above-threshold delta differences are stored and used for performing the multiplications (MACs). This technique significantly reduces computational complexity during inference and offers intermediate data compression opportunities. Our method does not require any training and can, therefore, be used directly in the existing pre-trained Transformer models. Moreover, no special and expensive hardware has to be developed as only comparisons are used in the algorithm. The evaluation is done on a pretrained Keyword Transformer model (KWT) \cite{kwtkws} using the Google Speech Commands Dataset (GSCD) \cite{gscd} with the focus on the accuracy-complexity trade-off. The results show that the number of computations can be reduced by $4.2x$ without losing any accuracy, and $7.5x$ while sacrificing 1\% of the baseline accuracy. Furthermore, the processing of the original MHSA block can be sped up by a factor of $\sim16$ while still achieving high accuracy of $\sim95\%$. Therefore, this work represents the next step to enable efficient inference of Transformers in low-power edge devices with the tinyML constraints.

\section{Related Work} \label{sec:relatedWork}
Different approaches have been used to reduce the computational complexity of the MHSA, such as cross-layer parameter sharing \cite{albert}, trimming individual weights \cite{weightsBERT} or removing encoders by distillation \cite{distillBERT, patientBERT, multitaskBERT, mobileBERT, simplerBERT}.
Recent research \cite{michel, differentiable, voita, mccarley} demonstrates that some attention heads can be eliminated without degrading the performance significantly. However, in order to obtain substantial computational savings and thus inference time gains, a considerable portion of heads would have to be discarded, inevitably leading to noticeable accuracy drops.\\
Other works focus on token pruning instead of removing redundant parameters. In \cite{powerbert}, redundant word-vectors are eliminated, outperforming previous distillation \cite{distillBERT, patientBERT} and head-pruning methods \cite{michel}. However, it requires training of a separate model for each efficiency constraint. This issue is resolved in \cite{lat} by adopting one-shot training that can be used for various inference scenarios, but the training process is complicated and involves multiple steps. Cascade pruning on both the tokens and heads is applied in \cite{spatten}, i.e., once a token and/or head is pruned, it is removed in all following layers. Nonetheless, this approach requires sorting of tokens and heads depending on their importance dynamically to select the top-k candidates, which needs specialized hardware. 
Similar to our work, recently published \cite{kim2021learned} also adopts a threshold-based pruning approach, which removes unimportant tokens as the input passes through the Transformer layers. However, this method requires a three-step training procedure to obtain a per-layer learned threshold, which again prevents to easily deploy the technique across a wide range of pre-trained networks. Most of the previous methods, moreover, only focus on optimizing Transformers for the natural language processing task.\\
The idea of threshold-based pruning using delta values for performing computations has already been explored for other types of DNNs, such as recurrent \cite{pmlrv70neil17a} and convolutional \cite{skipConvVideo} neural networks. 
However, incorporating a delta threshold in these networks results in significant memory overhead, as it requires storing intermediate states and activations. This issue is eliminated in our Delta Transformer, where almost no additional resources are required.\\

\vspace{-0.2cm}
\section{The Keyword Transformer}
\begin{figure}[t]
  \centering
  \includegraphics[width=1.0\linewidth]{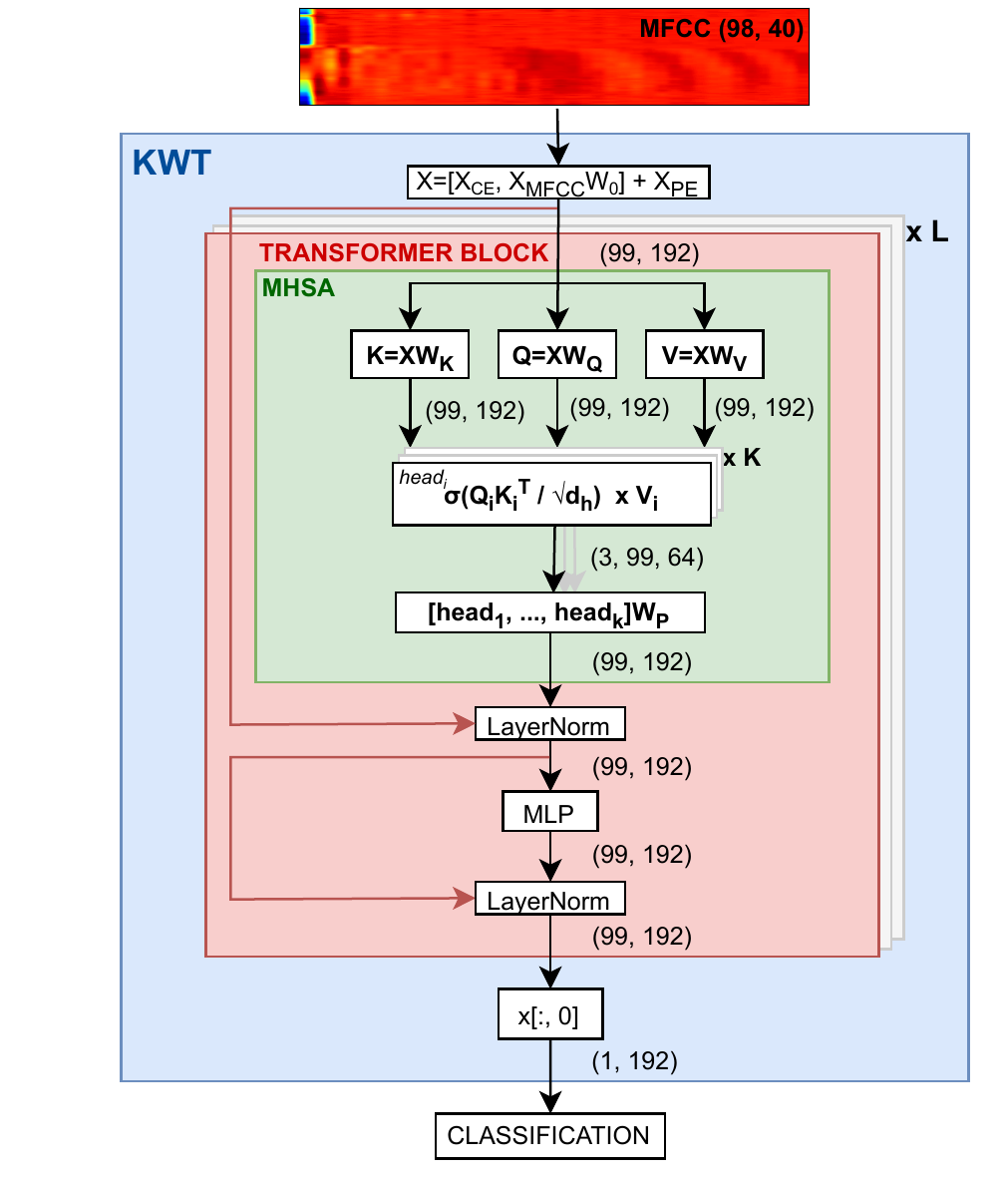}
  \caption{A high-level overview of the KWT model along with its dimensions. Red lines denote the residual connections.}
  \label{fig:TransformerEncoderOverview}
    \Description{A high-level overview of the KWT model along with its dimensions. Red lines denote the residual connections.}
    \vspace{-0.5cm}
\end{figure}

The typical Transformer encoder \cite{attentionisallyouneed} adopted in KWT consists of a stack of several identical Transformer blocks. Each Transformer block comprises of Multi-Head Self-Attention (MHSA), Multi-Layer Perceptron (MLP), layer normalizations, and residual connections as illustrated in Figure \ref{fig:TransformerEncoderOverview}. The key component in Transformers is the MHSA containing several attention mechanisms (heads) that can attend to different parts of the inputs in parallel.
We base our explanation on the KWT, proposed in \cite{kwtkws}. This model takes as an input the MFCC spectrogram of T non-overlapping patches $X_{MFCC} \in R^{T\,x\,F}$, with $t = 1,...,T$ and $f = 1,...,F$ corresponding to time windows and frequencies, respectively. This input is first mapped to a higher dimension $d$ using a linear projection matrix $W_0\in R^{F\,x\,d}$ along the frequency dimension, resulting in T tokens of dimension d. These are then concatenated with a learnable class embedding token $X_{CE} \in R^{1\,x\,d}$ representing a global feature for the spectrogram. Subsequently, a learnable positional embedding $X_{PE}\in R^{(T+1)\,x\,d}$ is added to form a final input to the Transformer encoder:
\vspace{-0.2cm}
\begin{equation}
\small
  X = [X_{CE};X_{MFCC}W_0] + X_{PE}
\end{equation}
The Transformer encoder multiplies the input $X$ with the projection matrices $W_Q, \,W_K,\, W_V \in R^{d\,x\,d}$, producing Query ($Q$), Key ($K$), and Value ($V$) input embedding matrices:
\begin{equation}
\vspace{-0.15cm}
\small
  Q = XW_Q; \qquad K = XW_K; \qquad V = XW_V
\vspace{0.01cm}
\end{equation}
The matrices are then divided into $k$ attention heads to perform the self-attention computations in parallel, where each of the heads $i=1, 2,..,k$ is given by:
\begin{equation}
\vspace{-0.2cm}
\small
  head_i = attention(Q_i, K_i, V_i) = softmax \left(\frac{Q_i(K_i)^T}{\sqrt{d_h}} \right) V_i
\end{equation}
The MHSA is defined as a concatenation of the attention heads, weighted by a projection matrix $W_P \in R^{kd_h \,x\,d}$, where $d_h=d/k$:
\begin{equation}
\small
  X_{MHSA}(Q, K, V) = [head_1, head_2,...,head_k]W_P
\end{equation}
The MHSA output is then added to the input $X$ with a residual connection and passed though the first layer normalization and the MLP block, followed by another addition of a residual input and second normalization:
\begin{equation}
\small
  X_{LN1} = LN(X_{MHSA} + X); \qquad  X_{LN2} = LN(X_{MLP} +  X_{LN1})
 \end{equation}
This structure is repeated $L$ times, denoting layers, to create an architecture of stacked Transformer layers.\\ 
In the KWT model, the MLP block is a two-layer feed-forward neural network using a GELU activation function after the first layer. The class embedding vector is extracted from the output of the last Transformer block to perform classification.\\
Three KWT models are proposed in the original work:  KWT-1 (607k parameters, $97.72\%\pm0.01$ accuracy), KWT-2 (2,394k parameters, $98.21\%\pm0.06$ accuracy), and KWT-3 (5,361k parameters, $98.54\%\pm0.17$ accuracy). We selected KWT-3 for our experiments, as it poses the biggest challenge as well as potential for compressing and reducing the computational complexity. The KWT-3 configuration is listed in Table \ref{tab:kwt3params}. 
\begin{table}[H]
\setlength{\tabcolsep}{0.5\tabcolsep}
  \caption{Configuration of the KWT-3 architecture}
  \label{tab:kwt3params}
    \begin{tabular}{llllll}
    \hline
    \textbf{Model} & dim $d$ & dim $d_{MLP}$ & heads $k$ & layers $L$ & \#params \\ \hline
    KWT-3          & 192 & 768     & 3     & 12     & 5,361k       \\ \hline
    \end{tabular}
\end{table}

\section{KWT model Analysis}
\begin{figure}[t]
  \centering
  \includegraphics[width=\linewidth]{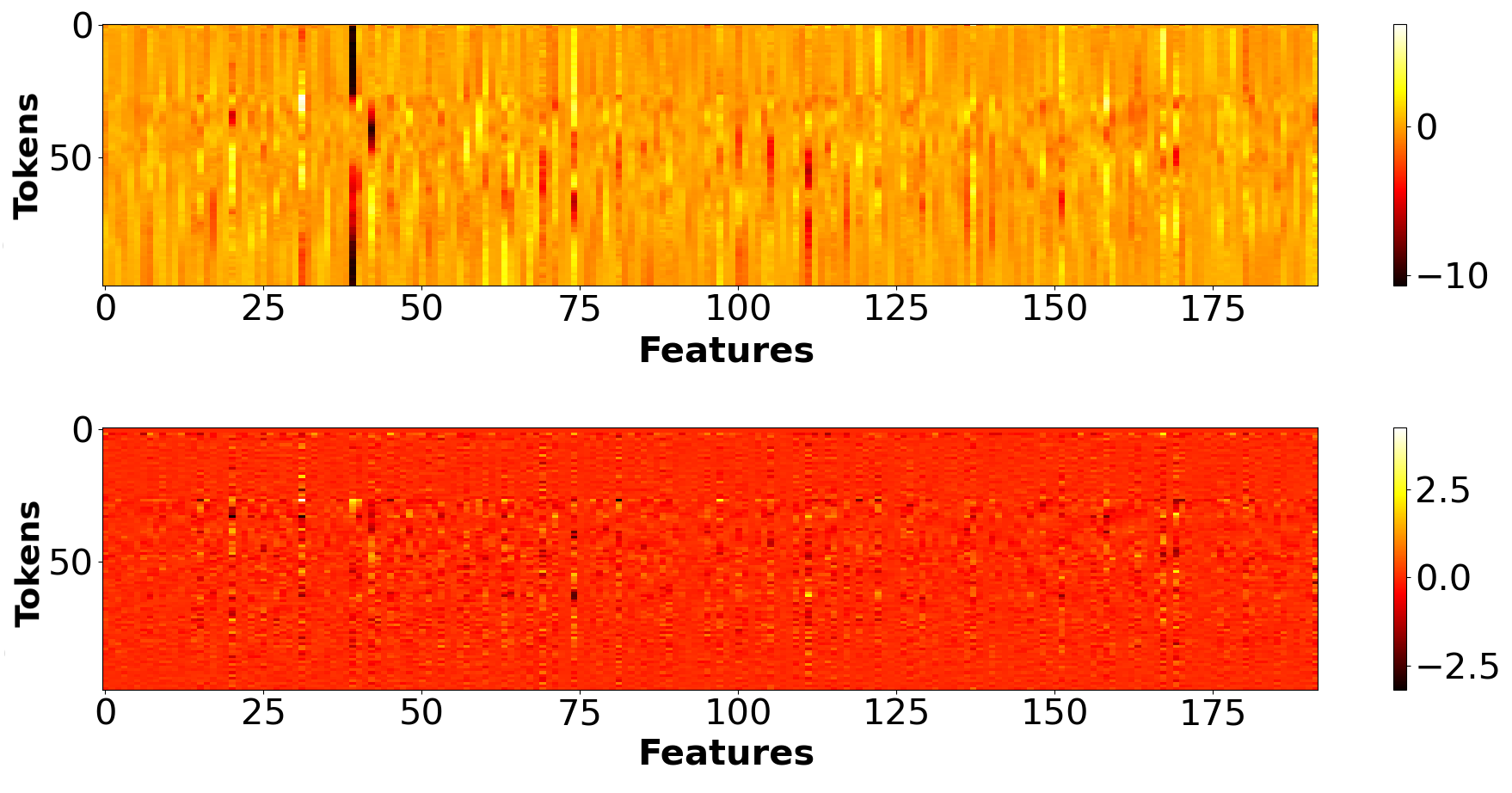}
  \caption{Input data to the 7th Transformer layer at the top along with its delta version at the bottom for keyword right.}
  \label{fig:inp7thLayer}
  \Description{Input data to the 7th Transformer layer at the top along with its delta version at the bottom for keyword right.}
\end{figure}

\begin{figure}[t]
  \centering
  \includegraphics[width=\linewidth]{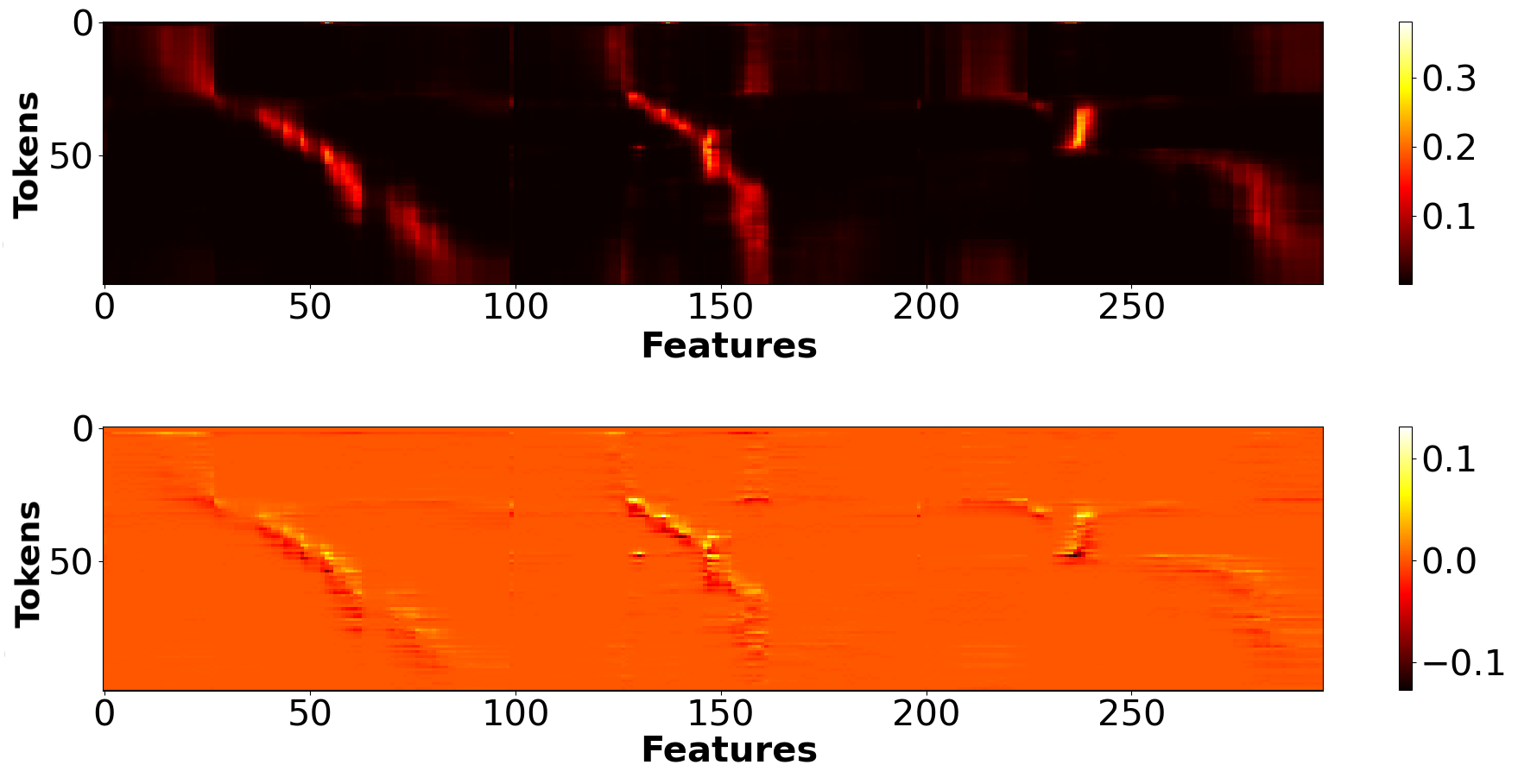}
  \caption{Softmax output of the 7th Transformer layer at the top along with its delta version at the bottom for the keyword $\mathbf{right}$. The figure illustrates three attention heads.}
  \label{fig:softmax7thLayer}
    \Description{Softmax output of the 7th Transformer layer at the top along with its delta version at the bottom for the keyword $\mathbf{right}$. The figure illustrates three attention heads.}
\end{figure}

\begin{figure}[t]
  \centering
  \includegraphics[width=\linewidth]{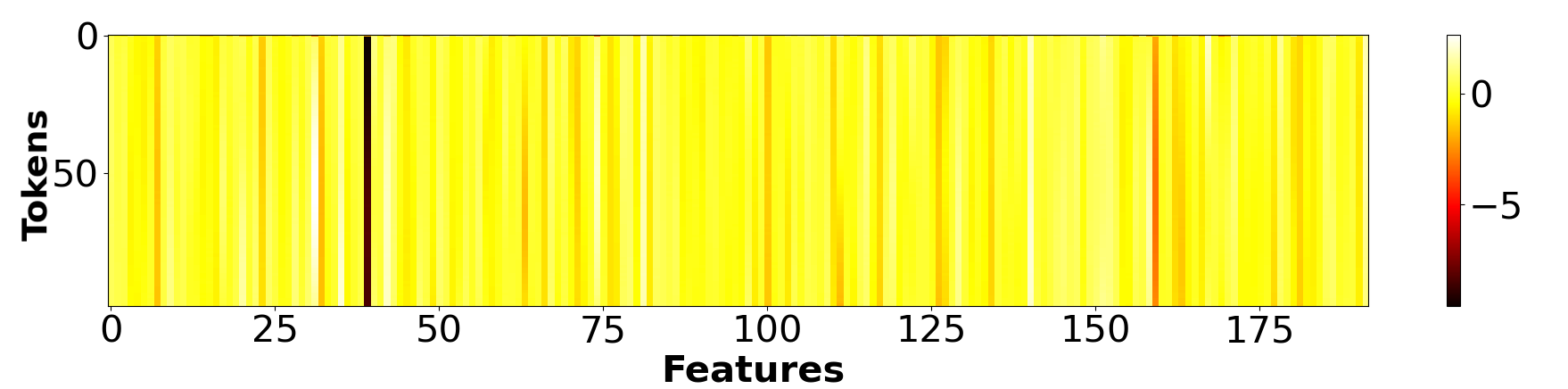}
  \caption{Input data to the 7th Transformer layer for $\mathbf{\_silence\_}$.}
  \label{fig:inp7thLayerSilence}
    \Description{Input data to the 7th Transformer layer for $\mathbf{\_silence\_}$.}
\end{figure}

The attention mechanism involves MACs of two matrices, resulting in $O(n^2)$ time and space complexity. However, as all tokens attend to each other, a certain level of redundancy is expected to be found in the system due to diffusion of information. Therefore, we analyze the KWT model on the GSCD to observe the degree of change across the tokens as they pass though the MHSA.
We feed multiple different keywords through the 12-layer KWT and inspect the MHSA inputs as well as intermediate results within the block. While considerable correlation across the tokens is expected for the initial input and intermediate results in the first layer, it is noteworthy to observe such behavior also in the MHSA of deeper layers, which is in line with cosine similarity measurements on word-vectors performed in \cite{powerbert}. Correlation is illustrated in Figure \ref{fig:inp7thLayer} showing the input $X$ (top) together with the difference between subsequent rows of this tensor (bottom), for the 7th layer of a keyword $right$. Figure \ref{fig:softmax7thLayer} repeats the same analysis for the softmax output of layer 7. It is clear that there is a significant amount of correlation between consecutive tokens, which opens up opportunities for data compression and/or computational data reuse. For example, $\sim84\%$ of the differences between corresponding features of subsequent tokens in $X$ are smaller than 1\% of the dynamic range of $X$ (7th layer). Such a tendency was observed for all voice-containing input sequences.
Moreover, when analyzing intermediate tensors from inputs of the $\_silence\_$ class, even larger data redundancy can be observed (Figure \ref{fig:inp7thLayerSilence}). 
It is clear that fully computing every single token would be a waste of computational and memory resources.
All these observations demonstrate that the amount of a significant change across the tokens constitutes only a small portion of the whole. Hence, introducing a threshold for recomputing could drastically decrease the computational load and inference time. Furthermore, exploiting sparsity across the tokens can also offer data compression. Therefore, we propose a delta algorithm that utilizes a threshold to discard insignificant values, further described in Section \ref{sec:deltaAlg}.

\section{Delta algorithm} \label{sec:deltaAlg}
\begin{figure}[t]
  \centering
  \includegraphics[width=\linewidth]{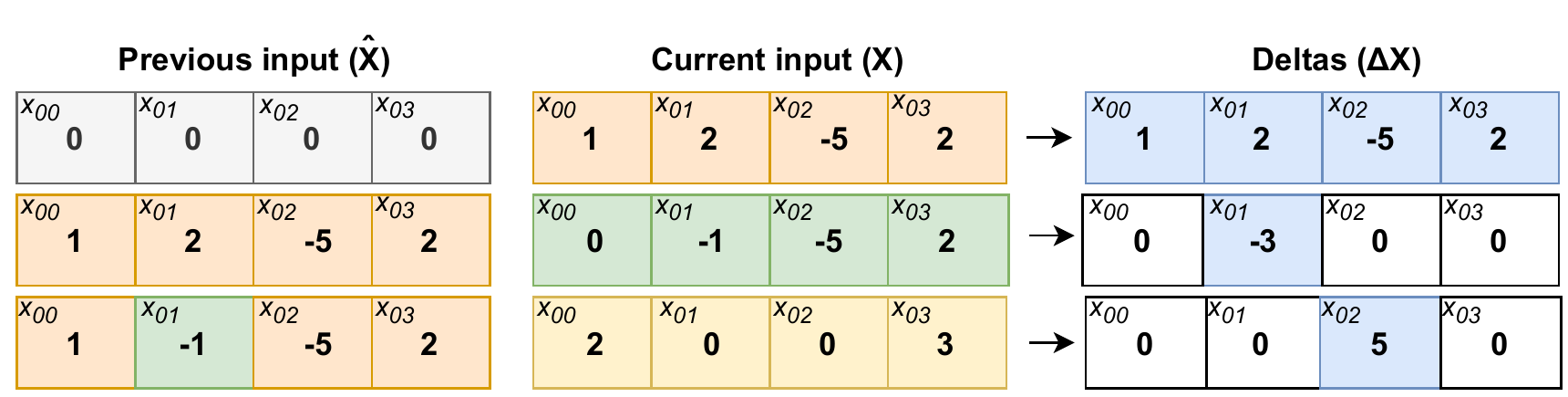}
  \caption{Delta algorithm example across three tokens with threshold $\mathbf{\theta=1.0}$. The top row corresponds to the first input vector that is always left untouched (no threshold).}
  \label{fig:deltaexample}
  \Description{Delta algorithm example across three tokens. The threshold is set to $\mathbf{\theta=1.0}$.}
\end{figure}

The objective of the delta algorithm is to transform a dense matrix-vector multiplication into a highly-sparse matrix-vector multiplication to reduce computational complexity and enable data compression, where only non-zero deltas are stored and used for computations.\\
The input $X$ always starts with the class embedding vector, followed by the first input vector. These two vectors (rows of the tensors) will always be left untouched throughout the complete MHSA pipeline. Every subsequent token after these will be represented by its delta value. This delta change $\Delta X (t)$ is calculated as the difference between the current input $X(t)$ and reference vector $\hat{X}(t-1)$. Only delta differences larger than a threshold $\theta$ are retained and used to update the reference vector $\hat{X}(t)$:
\begin{equation} \label{eq:xDelta}
\small
 \Delta X(t) = 
 \begin{cases}
      \small X(t) - \hat{X}(t-1) & \small \text{if $|X(t) - \hat{X}(t-1)|\,> \, \theta$} \\
      \small 0 & \small \text{otherwise}\\
    \end{cases} 
\end{equation}
\begin{equation} \label{eq:xPrevious}
\small
 \hat{X}(t) =  
 \begin{cases}
      \small X(t) & \small \text{if $|X(t) - \hat{X}(t-1)| \,> \, \theta$} \\ 
      \small \hat{X}(t-1) & \small \text{otherwise}\\
    \end{cases} 
\end{equation}
Where the $\hat{X}$ vector is initialized to 0s and updated once the first token arrives. Figure \ref{fig:deltaexample} visualizes this encoding over three tokens with $\theta=1.0$. The top row represents the first input vector that is left untouched (no delta algorithm applied). The orange and green colors in $\hat{X}$ show which values from the current input $X$ are propagated for the next token. White $\Delta X$ positions denote values of which magnitude equals to/is below $\theta$ and thus are skipped.\\
We apply the delta encoding of data at six different places in the MHSA: layer input $X$, matrices $K$ and $Q$, scaled $QK^T$, softmax output, and the attention head output. While the computations of delta values are the same everywhere, the subsequent operations with these deltas differ depending on whether i) a delta-encoded matrix is multiplied with a regular matrix, ii) two delta-encoded matrices are multiplied together, or iii) a non-linear $softmax$ function is applied. These three versions are described in the next subsections.

\subsection{Delta-regular matrix multiplication}
\begin{figure}[t]
  \centering
  \includegraphics[width=\linewidth]{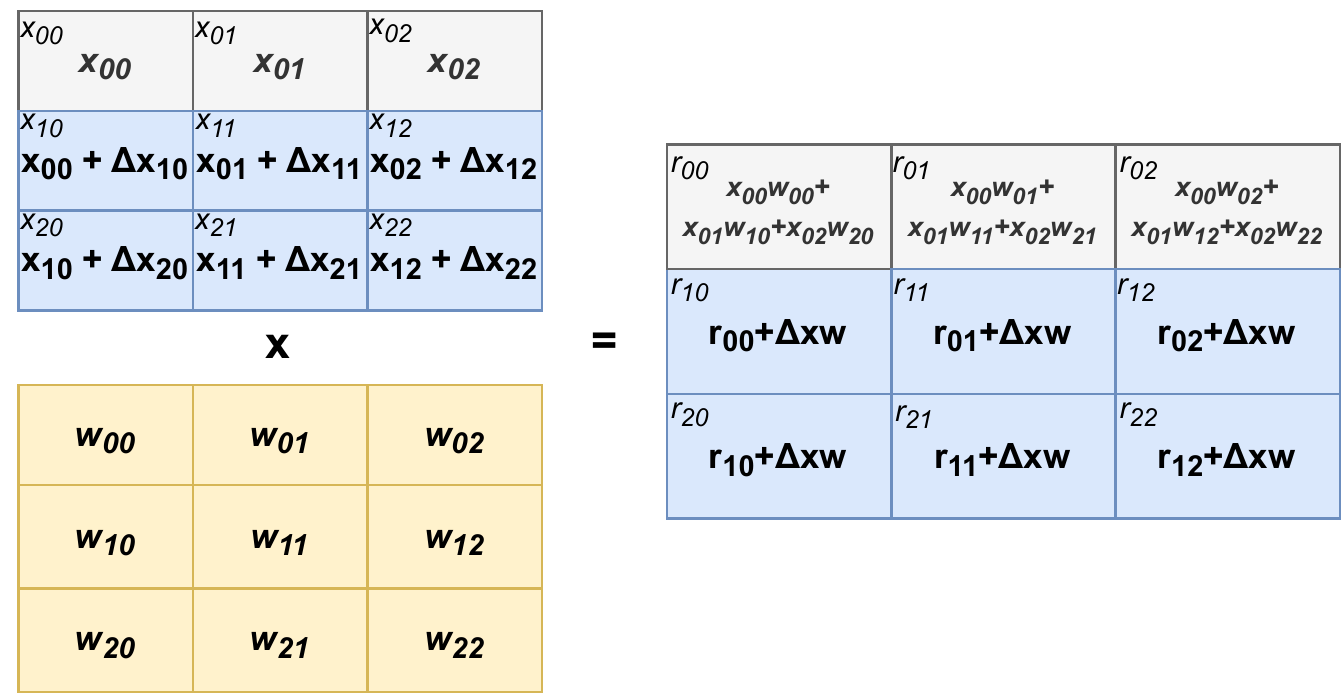}
  \caption{Baseline delta algorithm.}
  \label{fig:deltabaselinealg}
  \Description{Baseline delta algorithm.}
\end{figure}

Thanks to the delta representation, only non-zero $\Delta X$ are stored and used for multiplications as visualized in Figure \ref{fig:deltabaselinealg}. A weight matrix is denoted as $W$, and indices for $\Delta xw$ in the result matrix $R$ are excluded for clarity.  
The output $R(t)$ of the tensor operation can hence be computed by accumulating the result of the previous reference token $R(t-1)$ with the multiplication results of the weights with the delta values only. The updated $R(t)$ will then be the new baseline for the upcoming token:
\begin{equation} \label{eq:Xupdatemac}
\small
  R(t) = \Delta X(t)W + R(t-1)
\end{equation} 
  With $R(0)$ initialized to 0. These delta multiplications are used in $XW_Q$, $XW_K$, $XW_V$, $softmaxV$ and $[head_1,head_2,head_3]W_P$.
  
\subsection{Delta-delta matrix multiplication}    
\begin{figure}[t]
  \centering
  \includegraphics[width=\linewidth]{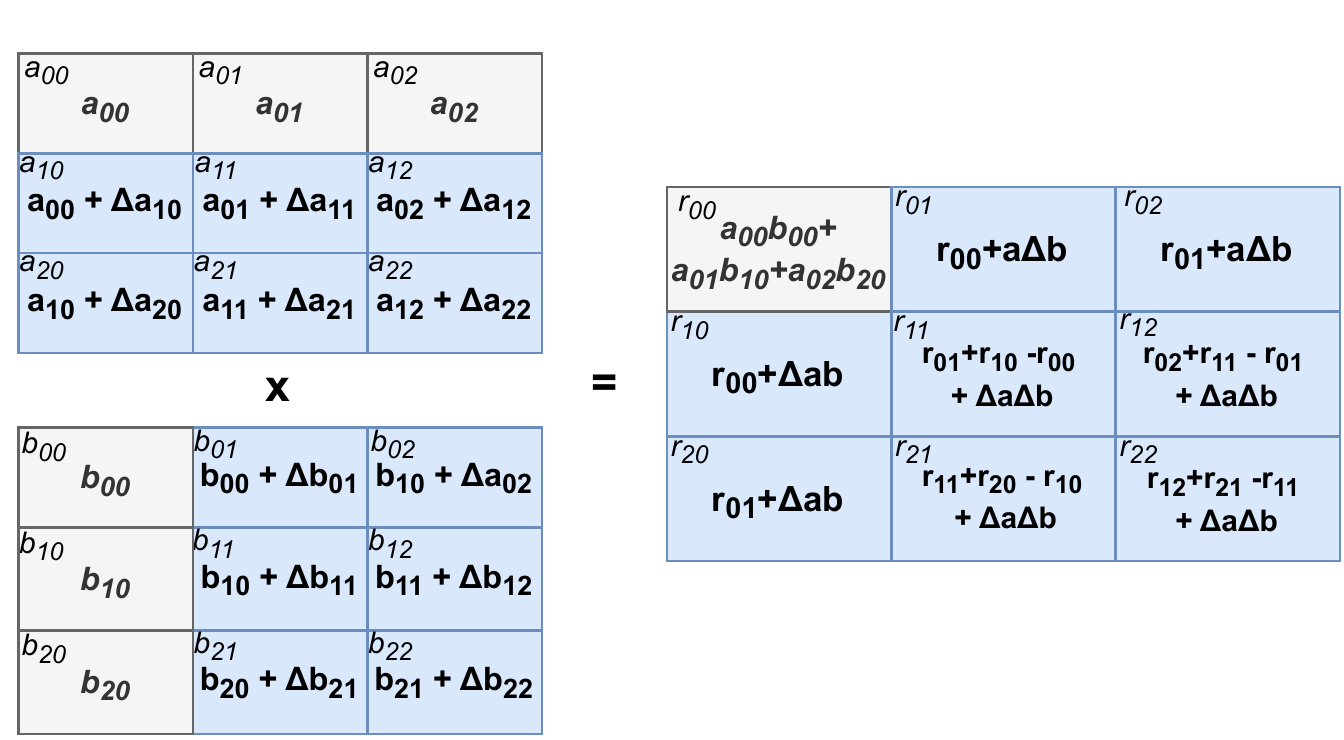}
  \caption{Delta algorithm for \boldmath{$QK^T$} represented with matrices $A$ and $B$.}
  \label{fig:deltaQKTalg}
  \Description{Delta algorithm for \boldmath{$QK^T$} represented with matrices $A$ and $B$.}
\end{figure}

\mv{\label{sec:qkt}}
As a result of the delta encoding, both $Q$ and $K$ will be expressed in their delta versions, and the multiplications will thus be slightly modified. This is described below and illustrated in Figure \ref{fig:deltaQKTalg} in a general form, with matrices $A$ and $B$ representing $Q$ and $K^T$, respectively. 
The multiplication of the first $A$ row with the first $B$ column is done as usually without using deltas:
\begin{equation} \label{eq:r00}
\small
  r_{00} = a_{00}b_{00} + a_{01}b_{10} + a_{02}b_{20}
\end{equation}
Then, the multiplication of the first $A$ row and second $B$ column exploits the delta approach in horizontal direction, where the $a_{00}b_{00} + a_{01}b_{10} + a_{02}b_{20}$ expression can be replaced with $r_{00}$ from eq. \ref{eq:r00} (marked with red):
\begin{flalign} \label{eq:r01}
\small
r_{01} &= a_{00}(b_{00} + \Delta b_{01}) + a_{01}(b_{10} + \Delta b_{11}) + a_{02}(b_{20} + \Delta b_{21}) &&\\\nonumber
&= a_{00}b_{00} + a_{00}\Delta b_{01}+ a_{01}b_{10}+ a_{01}\Delta b_{11}+ a_{02}b_{20} + a_{02}\Delta b_{21} &&\\\nonumber
&= \boldmath{\textcolor{red}{r_{00}}} + a_{00}\Delta b_{01}+ a_{01}\Delta b_{11}+ a_{02}\Delta b_{21}
\end{flalign}
Similarly, calculating results in the vertical direction for the rows of $A$ and first column of $B$ is given by:
\begin{flalign}\label{eq:r10}
\small
r_{10} &= b_{00}(a_{00} + \Delta a_{10})+ b_{10}(a_{01}+ \Delta a_{11})+ b_{20}(a_{02}+ \Delta a_{12}) &&\\\nonumber
&= b_{00}a_{00} + b_{00}\Delta a_{10}+ b_{10}a_{01}+ b_{10}\Delta a_{11}+ b_{20}a_{02}+ b_{20}\Delta a_{12} &&\\\nonumber
&= \boldmath{\textcolor{red}{r_{00}}} +\Delta a_{10}b_{00} + \Delta a_{11}b_{10}+ \Delta a_{12}b_{20}
\end{flalign}
An approach for multiplications for all the other positions is demonstrated on the second $A$ row and second $B$ column:
\begin{flalign} \label{eq:r11}
\small
r_{11} &= (\textcolor{blue}{a_{00}} + \textcolor{blue}{\Delta a_{10}})
          (\textcolor{blue}{b_{00}} + \textcolor{blue}{\Delta b_{01}})
          + (\textcolor{green}{a_{01}}+ \textcolor{green}{\Delta a_{11}})
          (\textcolor{green}{b_{10}}+ \textcolor{green}{\Delta b_{11}}) &&\\\nonumber
        &+ (\textcolor{orange}{a_{02}}+ \textcolor{orange}{\Delta a_{12}})
        (\textcolor{orange}{b_{20}} + \textcolor{orange}{\Delta b_{21}}) &&\\\nonumber
&= \textcolor{blue}{a_{00}b_{00}} + \textcolor{blue}{a_{00} \Delta b_{01}} + \textcolor{blue}{\Delta a_{10}b_{00}} + \textcolor{blue}{\Delta a_{10}\Delta b_{01}} &&\\\nonumber
&+ \textcolor{green}{a_{01}b_{10}} + \textcolor{green}{a_{01} \Delta b_{11}} + \textcolor{green}{\Delta a_{11}b_{10}} + \textcolor{green}{\Delta a_{11}\Delta b_{11}} &&\\\nonumber
&+ \textcolor{orange}{a_{02}b_{20}} + \textcolor{orange}{a_{02} \Delta b_{21}} + \textcolor{orange}{\Delta a_{12}b_{20}} + \textcolor{orange}{\Delta a_{12}\Delta b_{21}} &&\\\nonumber
&= \textcolor{red}{r_{01}} + \textcolor{red}{r_{10}} - \textcolor{red}{r_{00}} + 
\textcolor{blue}{\Delta a_{10}\Delta b_{01}} + \textcolor{green}{\Delta a_{11}\Delta b_{11}} + \textcolor{orange}{\Delta a_{12}\Delta b_{21}}
\end{flalign}

Where different colors mark each of the three multiplications. Simplifying parenthesis shows that the expressions not involving any deltas can be substituted with $r_{00}$. Next, the terms with $\Delta b$ are replaced with $r_{01}$, while those containing $\Delta a$ with $r_{10}$. Since $r_{00}$, $r_{01}$, and $r_{10}$ have already been computed in previous timesteps, we only need to do the (sparse) delta multiplications themselves and subtract the $r_{00}$ result as it is present in both $r_{01}$ and $r_{10}$. These steps are then applied to all the other slots as shown in Figure \ref{fig:deltaQKTalg}.\\
\vspace{-0.3cm}
\subsection{Delta for softmax} 
Delta algorithm cannot be directly applied for softmax as this function introduces a non-linearity to the system:
\begin{equation} \label{eq:softmax}
\small
  softmax(r)_{i} = \frac{exp(r_i)}{\sum_{j}^{}exp(r_j)}
\end{equation}
We will have to introduce a scaling factor to correct the softmax computations. As done earlier, we will again start by performing unaltered processing of the initial row  $  r_{0} = [r_{00} \, r_{01} \, r_{02}]$ (class embedding excluded for clarity) with a regular softmax function:
\begin{equation} \label{eq:deltasoftmaxR1}
\small
  softmax(r)_{0} = \frac{[exp(r_{00}) \,exp(r_{01}) \, exp(r_{02})]}
  {\sum [exp(r_{00}) \, exp(r_{01}) \, exp(r_{02})]} 
\end{equation}
The next row of the scaled input $QK^T$ is already expressed with deltas:
\begin{equation} \label{eq:R2deltas}
\small
  r_{1} = [\Delta r_{10} \, \Delta r_{11} \, \Delta r_{12}]
\end{equation}
The $r_1$ nominator $NOM_{r_1}$ for softmax is thus given by:
\begin{equation} \label{eq:R2nom}
\small
  NOM_{r_1} = [exp(\Delta r_{10}) \,exp(\Delta r_{11}) \, exp(\Delta r_{12})]
\end{equation}
While the denominator $DENOM_{r_1}$ as: 
\begin{equation} \label{eq:R2denom}
\small
  DENOM_{r_1} = \frac{\sum [exp(r_{00}+ \Delta r_{10}) \, exp(r_{01}+\Delta r_{11}) \, exp(r_{02}+\Delta r_{12})]} 
  {\sum [exp(r_{00}) \, exp(r_{01}) \, exp(r_{02})]} 
\end{equation}
Finally, a scaling factor for each of the values to correct the softmax result is: 
\begin{equation} \label{eq:R2scalingfactor}
\small
  SF_{r_1} = softmax(r)_{1} \, \frac{NOM_{r_1}}{DENOM_{r_1}}
\end{equation}

\subsection{Computational savings} \label{subsec:deltaAlgSavings}
To assess the potential computational savings for the Delta KWT, we differentiate between the two main sublayers: i) MHSA, and ii) MLP. 
The MLP block consists of two fully connected layers with weight matrices of dimensions (192,768) and (768,192), respectively. Without any delta modification, $\sim$39\% of the multiplication of the original KWT can be found in the MHSA and $\sim$61\% in the MLP. Although MLP is the prevailing module in this specific scenario, its complexity does not grow quadratically with the input sequence length. Moreover, there are many well-established compression techniques available, some of them presented in Section \ref{sec:relatedWork}. Hence, pruning of the MLP is out of the scope of our work, and it is only stated for completeness. The MHSA multiplication operations can be further split into $XW_K$, $XW_Q$, $XW_V$ ($\sim$59.63\%), $QK^T$ ($\sim$10.25\%), $softmax(QK^T)V$ ($\sim$10.25\%), and final projection with attention heads $[head_1,head_2,head_3]W_P$ ($\sim$19.88\%). The KWT model offers an optimization in the last layer. As shown in Figure \ref{fig:TransformerEncoderOverview}, only the class embedding token is used for the final prediction, making the rest of the tokens within the sequence unused. This dependency can be tracked up to $QK^T$. The MAC savings in last layer are thus worth $59.64\%$, always making the total savings at least $4.97\%$ for the whole KWT without losing any accuracy.\\  
Maximum possible computational savings, i.e., cases when only the class embedding and first vector are computed since all deltas are 0, are stated below for each of the MHSA parts. For simplicity, all the terms use matrices $A$ and $B$, and $row$ and $col$ for dimensions.\\
Savings for $XW_K$, $XW_Q$, and $XW_V$ for each of the first 11 layers are:
\begin{equation} \label{eq:saveInp11}
\small
  l_{0-10} = 1 - \frac{( colA \, x\, 2) \,x\, colB\, x\, 3}{(colA \,x\, colB \,x\, rowA)\, x\, 3} = \sim97.98\%
\end{equation}
Where $A=(99,192)$ and $B=(192,192)$. Computations for $XW_Q$ in the last layer are expressed as:
\begin{equation} \label{eq:saveInp12}
\small
  l_{11} = 1 - \frac{(colA \, x\, 2)\, x\, colB\, x \, 2  + \,colA \,x\,\, colB}{(colA \,x \,colB \,x \,rowA)\,x\, 3} = \sim98.32\%
\end{equation}
Savings for $QK^T$:
\begin{equation} \label{eq:saveQKt11}
\small
  l_{0-10} = 1 - \frac{( colA\, x\, 2 \,x\, 2)\, x \,heads}{(colA \,x\, colB\, x \,rowA) \,x\, heads} = \sim99.96\%
\end{equation}
\begin{equation} \label{eq:saveQKt12}
\small
  l_{11} = 1 - \frac{(colA\, x\, 2)\, x\, heads}{(colA\, x\, colB\, x\, rowA) \,x \,heads} = \sim99.98\%
\end{equation}
Where $A=(99,64)$ and $B=(64,99)$. Savings for $softmax(QK^T)V$:
\begin{equation} \label{eq:savesoftmaxV11}
\small
  l_{0-10} = 1 - \frac{(colA \, x\, 2\, x \,colB) \,x\, heads}{(colA\, x\, colB\, x\, rowA) \,x \,heads} = \sim97.98\%
\end{equation}
\begin{equation} \label{eq:savesoftmaxV12}
\small
 l_{11} = 1 - \frac{(colA \, x \,colB) \,x\, heads}{(colA\, x\, colB\, x\, rowA) \,x \,heads} = \sim98.99\%
\end{equation}
Where $A=(99,99)$ and $B=(99,64)$. Finally, the projection with attention heads:
\begin{equation} \label{eq:savechV11}
\small
  l_{0-10} = 1 - \frac{(colA \, x\, 2) \,x\, colB}{colA \,x\, colB \,x\, rowA} = \sim97.98\%
\end{equation}
\begin{equation} \label{eq:savechV12}
\small
  l_{11} = 1 - \frac{colA \,x\, colB}{colA \,x\, colB \,x\, rowA} = \sim98.99\%
\end{equation}
Where $A=(99,192)$ and $B=(192,192)$. \\
Of course, the savings estimated above only hold for the extreme case, which means that either a) all tokens are perfectly correlated, or b) very large thresholds are used, resulting in significant accuracy degradation. Section \ref{sec:ResDis} will therefore analyze the complete accuracy-complexity trade-off for real data sequences. 

\subsection{Resources}
The proposed delta approach neither requires expensive hardware nor comes with a large memory overhead. Only a single token has to be stored as a reference whenever the delta method is used. The softmax delta version additionally needs to keep the sum of exp from one timestep to another. In terms of computations, an additional division is needed when calculating scaling factors, along with multiplications with scaling factors for features within a token.\\
The downside of our method is compute and data irregularity due to the algorithm's \textit{unstructured} pruning. However, there are many techniques proposed in literature such as \cite{unstrPrunAccel} on how to handle this challenge.

\section{Experimental setup}
The GSCD v2 \cite{gscd} is used to evaluate our method as well as the original KWT performance. The dataset contains 105,000 1-second audio snippets of 35 different words sampled at 16 kHz. The model classifies 4,800 keywords from a test set into one of the 12 categories: ”up”, ”down”, ”left”, ”right”, ”yes”, ”no”, ”on”, ”off”, ”go”, and ”stop”, ”\_silence\_” and ”\_unknown\_”.\\
To assess the impact of the thresholds for the different parts of the MHSA on accuracy and model complexity, we executed threshold sweeps on a subset of 100 keywords (6-12 words from each category). While the thresholds might be different for each delta encoding within the MHSA block, they are the same across every Transformer layer. This means that MHSA in the first layer uses the same thresholds as MHSAs in other layers. From these sweeps, the thresholds leading to a Pareto-optimal accuracy-computations trade-off are used in a full run with all 4,800 keywords. We focused on those configurations that yielded at least 94\% accuracy. Since the thresholds are first determined on a subset of the complete dataset, it was expected to obtain variations in the results when performing the test on the full dataset. Additional finetuning, i.e., threshold adjusting, was done and the results are presented and discussed in Section \ref{sec:ResDis}.\\
\vspace{-0.3cm}
\section{Results and Discussion} \label{sec:ResDis}
\begin{figure}[t]
  \centering
  \includegraphics[width=\linewidth]{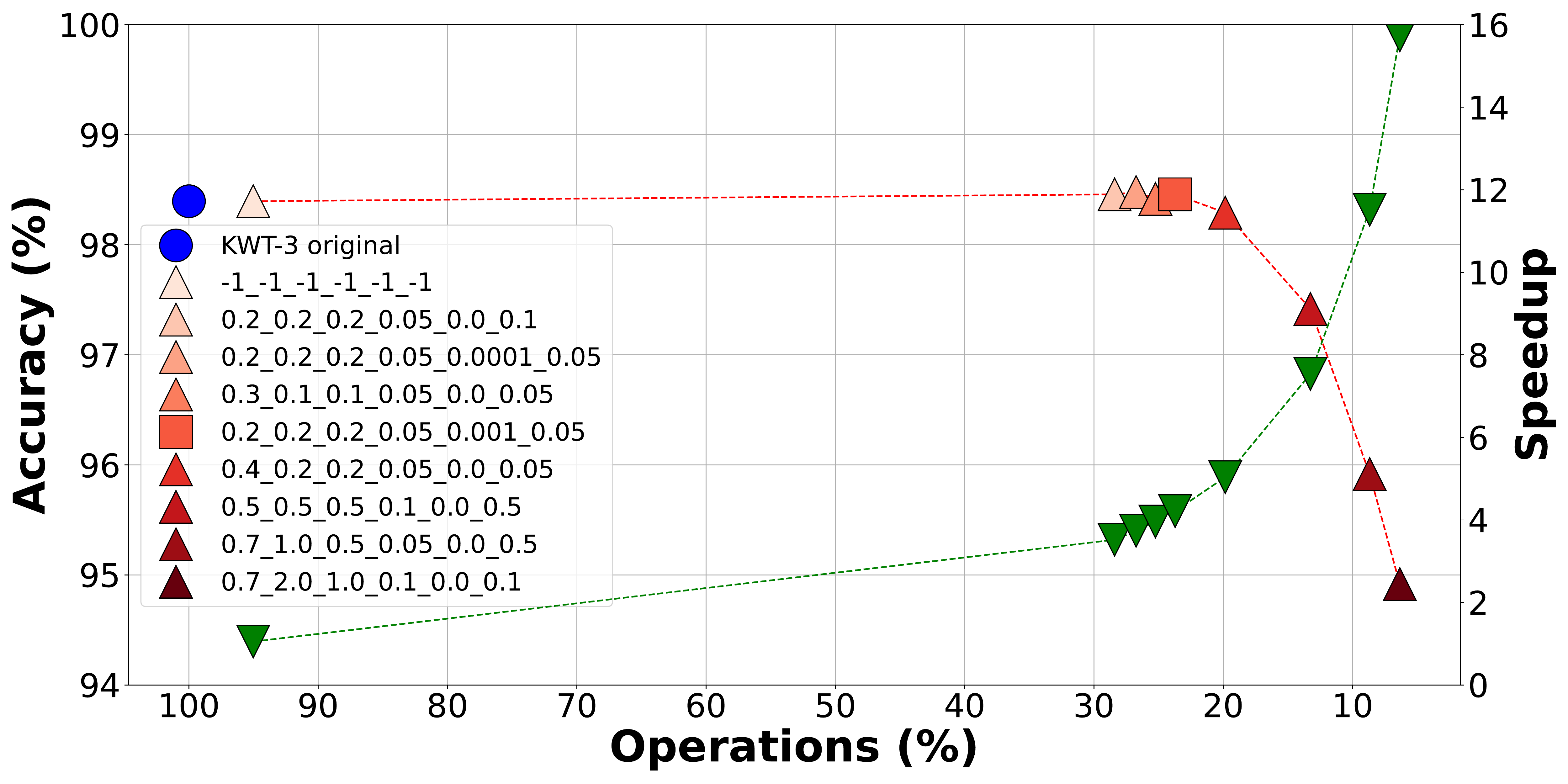}
  \caption{The results of running the original and the delta version of the KWT model. X-axis represents MACs, while the left and right y-axis correspond to the accuracy and speedup, respectively. Each of the red-shaded triangles (and a square) in the legend is annotated with thresholds used during the experiment in order: $\mathbf{\theta_X}$, $\mathbf{\theta_Q}$, $\mathbf{\theta_K}$, $\mathbf{\theta_{QK^T}}$, $\mathbf{\theta_{softmax}}$, and $\mathbf{\theta_{head_{1..k}}}$}
  \label{fig:finalRes}
  \Description{The results of running the original as well as the delta version of the KWT model. X-axis represents MACs, while the left and right y-axis correspond to the accuracy and speedup, respectively. Each of the red-shaded triangles (and a square) in the legend is annotated with thresholds used during the experiment in order: $\mathbf{\theta_X}$, $\mathbf{\theta_Q}$, $\mathbf{\theta_K}$, $\mathbf{\theta_{QK^T}}$, $\mathbf{\theta_{softmax}}$, and $\mathbf{\theta_{head_{1..k}}}$}
\end{figure}

\begin{figure}[t]
    \centering
    \subfloat[Delta inputs]{{\includegraphics[width=0.45\linewidth]{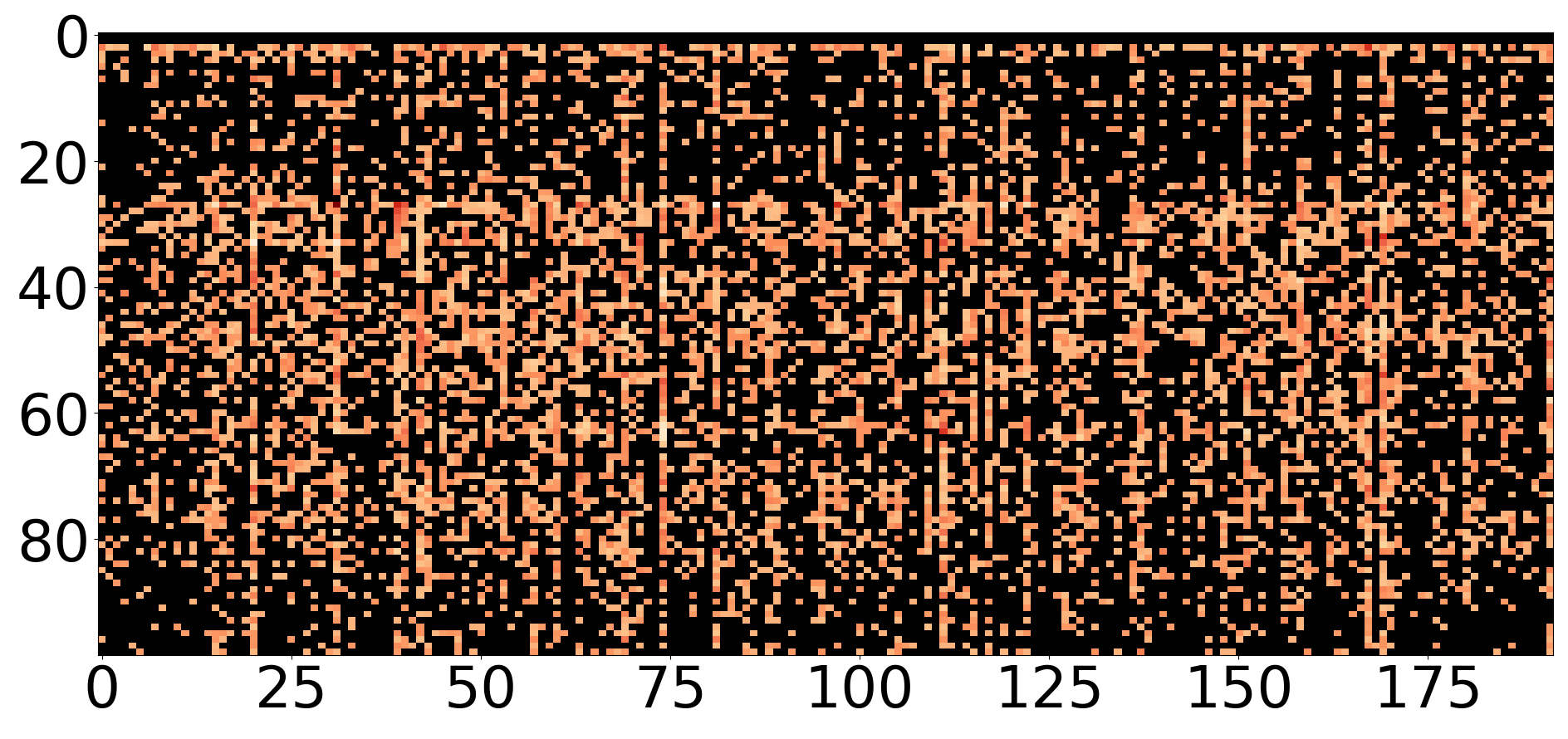} }}
    \qquad
    \subfloat[Delta softmax outputs (three attention heads) ]{{\includegraphics[width=0.45\linewidth]{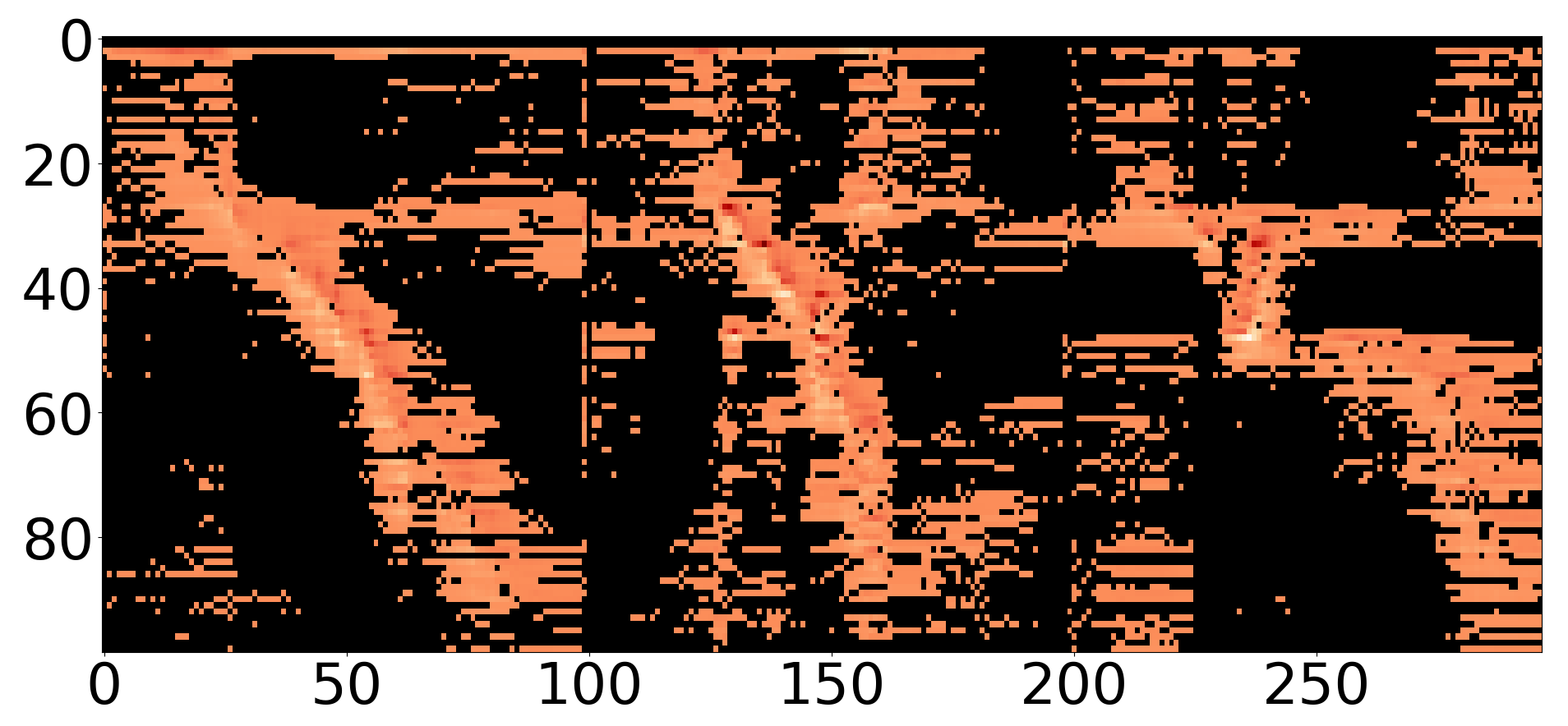} }}
    \caption{Deltas for a) inputs and b) softmax outputs for the 7th Transformer layer of the keyword $\mathbf{right}$. Black color marks 0s.}
    \label{fig:deltas}
    \Description{Deltas for a) inputs and b) softmax outputs for the 7th Transformer layer of the keyword $\mathbf{right}$. Black color marks 0s.}
\end{figure}

\begin{figure}[t]
  \centering
  \includegraphics[width=0.6\linewidth]{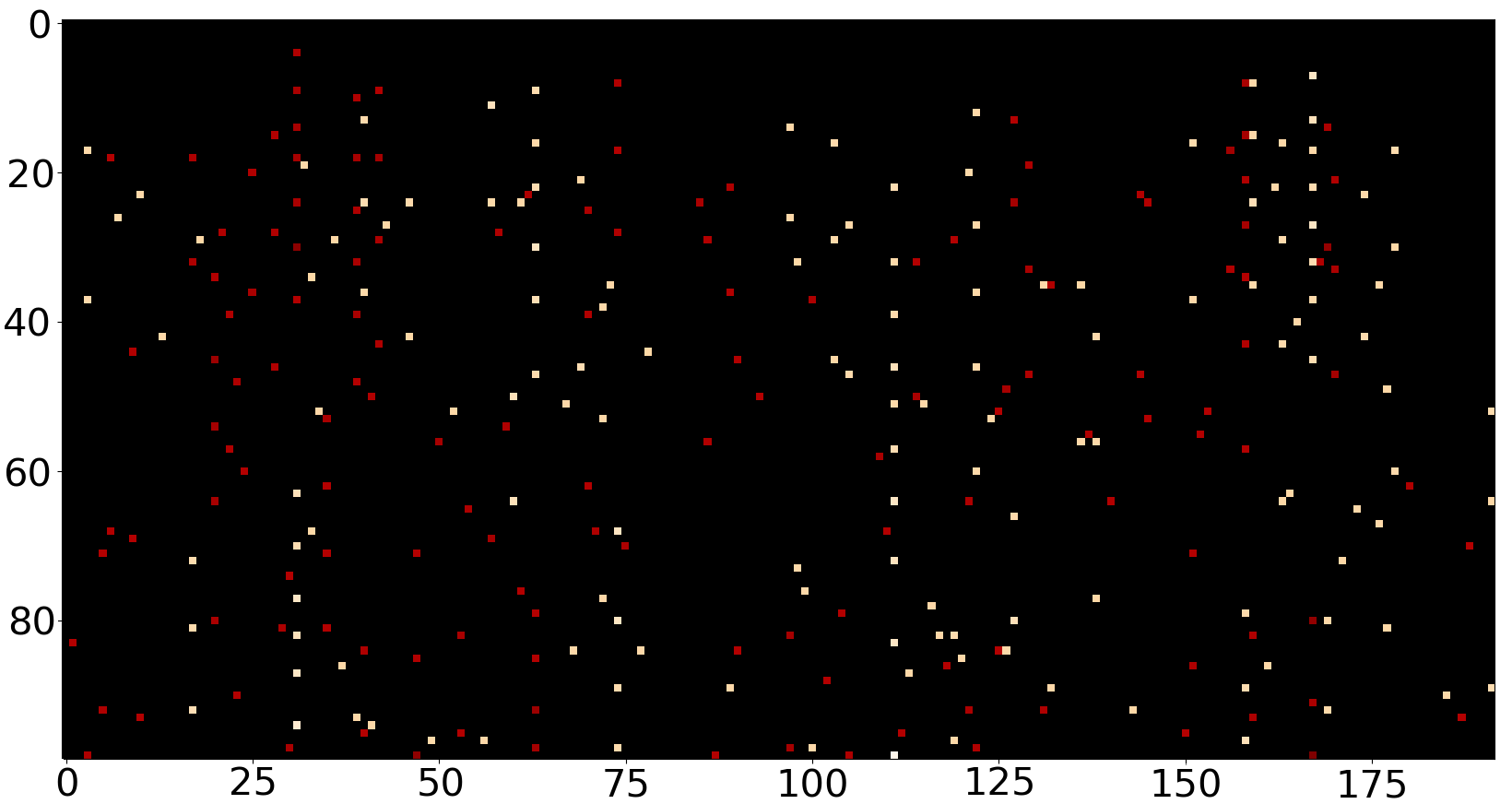}
  \caption{Deltas for inputs to the 7th Transformer layer for $\mathbf{\_silence\_}$. Black color marks 0s.}
  \label{fig:inpDeltasSilence}
  \Description{Deltas for inputs to the 7th Transformer layer for $\mathbf{\_silence\_}$. Black color marks 0s.}
\end{figure}

The Pareto-optimal results evaluated on all 4,800 audio files are shown in Figure \ref{fig:finalRes}, where the delta configurations are provided in the legend.
The x- and the left y-axis show a percentage of executed MACs averaged across the layers and achieved accuracy, respectively. The second y-axis represents a speedup factor derived from the amount of MACs. The blue circle corresponds to the original KWT-3 model that achieves $\sim98.4\%$ accuracy with 100\% MACs.   
\begin{table}[t]
\setlength{\tabcolsep}{0.4\tabcolsep}
\renewcommand{\arraystretch}{0.5}
 \caption{Percentage of executed MACs averaged across the layers for one instance of each keyword category. The configuration is: $\mathbf{\theta_X=0.2}$, $\mathbf{\theta_Q=0.2}$, $\mathbf{\theta_K=0.2}$, $\mathbf{\theta_{QK^T}=0.05}$, $\mathbf{\theta_{softmax}=0.001}$, and $\mathbf{\theta_{head_{1..k}}=0.05}$.}
   \label{tab:keywordOps}
\begin{tabular}{|c|c|c|c|c|c|}
\hline
\textbf{Keyword}     & {$\mathbf{XW_{Q, K, V}}$} & {$\mathbf{QK^T}$} & \textbf{$\mathbf{softmaxV}$} & $\mathbf{head_{1..k}W_P}$ & \textbf{Total} \\ \hline
\textbf{\_silence\_} &    3.02                 &  0.08               &    2.4              &  2.02                &  2.46                   \\ \hline
\textbf{\_unknown\_} &     35.97                &  7.68                &    26.95              & 16.93                 & 28.36                    \\ \hline
\textbf{yes}         &  31.82                   &    6.26              &    24.06              & 16.31                 & 25.32                    \\ \hline
\textbf{no}          &  36.98                   &  8.93                &  25.42                & 14.27                  &  28.41                   \\ \hline
\textbf{up}          & 33.48                    &   6.28               &       23.64           & 13.32                 &    25.68                 \\ \hline
\textbf{down}        &  29.88                   &    5.08              & 22.38                 & 14.24                 &   23.46                  \\ \hline
\textbf{left}        &  38.73                   &  10.09                &  26.95                & 16.97                  &  30.26                   \\ \hline
\textbf{right}       & 33.52               &   6.68               &     25.65             &  16.55                 &  26.59                   \\ \hline
\textbf{on}          &  31.13                   &    5.64              &   21.21               & 13.02                 &    23.9                 \\ \hline
\textbf{off}         &  39.5                   &    10.17              &    26.68              & 15.11                 &          30.33           \\ \hline
\textbf{stop}        &  32.39                  &    6.15              &   23.36               &  13.72                &    25.06                 \\ \hline
\textbf{go}          &   33.37                  &    6.57              &    23.26              & 14.65                 &       25.87              \\ \hline
\end{tabular}
\end{table}
The red and green triangles represent our delta KWT-3 model with regard to accuracy and speedup, respectively.
The inference time gains for the MHSA range from $\sim1.05x$ to $\sim16x$, and there is no accuracy degradation down to $\sim23.7\%$ MACs (4.2x speedup). Moreover, some of the configurations even slightly outperform the original KWT-3 (98.46\%, 98.48\%, and 98.42\%). Decreasing the accuracy by only 0.1\% results in further speedup of $5x$. Moreover, if the accuracy requirements can be relaxed by 1-4\%, the MHSA inference becomes faster by $7.5-15.7x$, which translates to 86.73-93.65\% of skipped MACs.
Table \ref{tab:keywordOps} shows the \% of executed MHSA operations for one instance of each keyword category, averaged across the layers. The configuration (0.2\_0.2\_0.2\_0.05\_0.001\_0.05) used to obtain the results is represented with a square in Figure \ref{fig:finalRes}. Although the MAC percentage naturally fluctuates for keywords within the same group, the objective of the table is to provide a general overview of how much operations are approximately performed in each of the parts. We can observe that $\sim60-70\%$ of $XW_{Q,K,V}$, $90-95\%$ of $QK^T$, $73-79\%$ of $softmaxV$, and $83-87\%$ of $head_{1..k}W_P$ are discarded, which sums up to $70-77\%$ of skipped operations for the entire model.
To visualize the savings, Figure \ref{fig:deltas} shows the delta values of the input data $X$ and the softmax output of the 7th layer of a keyword $right$ (same instance as used in Table \ref{tab:keywordOps}).
One special case are the instances from the $\_silence\_$ class, that have the amount of discarded computations very close to the theoretical maximum defined in Section \ref{subsec:deltaAlgSavings}. Figure \ref{fig:inpDeltasSilence} shows the $\_silence\_$ input, for which only a small fraction of the deltas are non-zero, resulting in $97-99.9\%$ of skipped operations.\\
A potential future improvement involves applying deltas on the input embedding matrix $V$. Although these cannot be exploited in multiplications with the softmax output due to the direction of computations (softmax output compensates for it), it would still contribute to $V$'s data compression.
Future work also explores the most optimal thresholds for each of the layers individually. This might further optimize the point where the accuracy starts dropping since a varying number of MACs is executed within each of the 12 layers.

\section{Conclusion}
This paper introduced a dynamic threshold-based pruning technique that drastically reduces MAC operations during inference. It was demonstrated on a keyword spotting task on the GSCD, where $\sim80\%$ of operations in the MHSA can be discarded without degrading the accuracy. If the accuracy requirements can be slightly relaxed, a speedup factor of $\sim5-16x$ is achieved. Our method thus helps to considerably decrease the computational complexity and enable significant data compression. The proposed technique can be exploited to enable an ultra-low power wake-up word detection front-end, that triggers a more powerful detector once a keyword is recognized.
More generally, this work represents a stepping stone towards enabling the execution of Transformers on low-power devices. 


\bibliographystyle{ACM-Reference-Format}
\bibliography{acmart}

\end{document}